\documentclass[a4paper,12pt]{elsarticle}

\usepackage[T2A]{fontenc}
\usepackage[utf8]{inputenc}
\usepackage[russian,english]{babel}

\usepackage{amsmath}
\usepackage{amsfonts}
\usepackage{amssymb}
\usepackage{amsthm}

\usepackage{hhline}

\newtheorem{theorem}{Theorem}[section]

\theoremstyle{definition}
\newtheorem{example}{Example}
\theoremstyle{remark}

\usepackage{bookmark}

\usepackage{hyperref} 

\usepackage{graphicx} 
\usepackage{amssymb,amsfonts,amsmath,amsthm} 
\usepackage[usenames,dvipsnames]{color} 
\usepackage{makecell}
\usepackage{multirow} 
\usepackage{ulem} 
\usepackage[table,xcdraw]{xcolor}
\usepackage{wrapfig}
\usepackage{pgfplots}
\pgfplotsset{compat=newest}
\usepackage{geometry}
\usepackage{cancel} 

\usepackage[matrix,arrow,curve]{xy}

\begin{document}

\begin{frontmatter}

\title{Rosenblatt's first theorem and frugality of deep learning}

\author[NNU,ICM]{A. N. Kirdin\corref{cor1}}
\ead{kirdinalexandergp@gmail.com}
\author[NNU]{S. V. Sidorov}
\ead{sesidorov@yandex.ru}
\author[NNU]{N. Y. Zolotykh}
\ead{nikolai.zolotykh@gmail.com}
\address[NNU]{Lobachevsky University, Nizhni Novgorod, Russia}
\address[ICM]{Institute for Computational Modelling, Russian Academy of Sciences, Siberian Branch, Krasnoyarsk, Russia}
 \cortext[cor1]{Corresponding author}

\begin{abstract}
First Rosenblatt's theorem about omnipotence of shallow networks states that elementary perceptrons can solve any classification problem if there are no discrepancies in the training set. Minsky and Papert considered elementary perceptrons with restrictions on the neural inputs: a bounded number of connections or a relatively small diameter of the receptive field for each neuron at the hidden layer. They proved that under these constraints, an elementary perceptron cannot solve some problems, such as the connectivity of input images or the parity of pixels in them. In this note, we demonstrated first Rosenblatt's theorem at work, showed how an elementary perceptron can solve a version of the travel maze problem, and analysed the complexity of that solution. We constructed also a deep network algorithm for the same problem.  It  is much more efficient. The  shallow network uses an exponentially large number of neurons on the hidden layer (Rosenblatt's $A$-elements), whereas for the deep network  the second order polynomial  complexity is sufficient. We demonstrated that for the same complex problem deep network can be much smaller and reveal a heuristic behind this effect. 
\end{abstract}
\end{frontmatter}

\section{Introduction}

\begin{figure}[h]\centering
\includegraphics[width=0.6\textwidth]{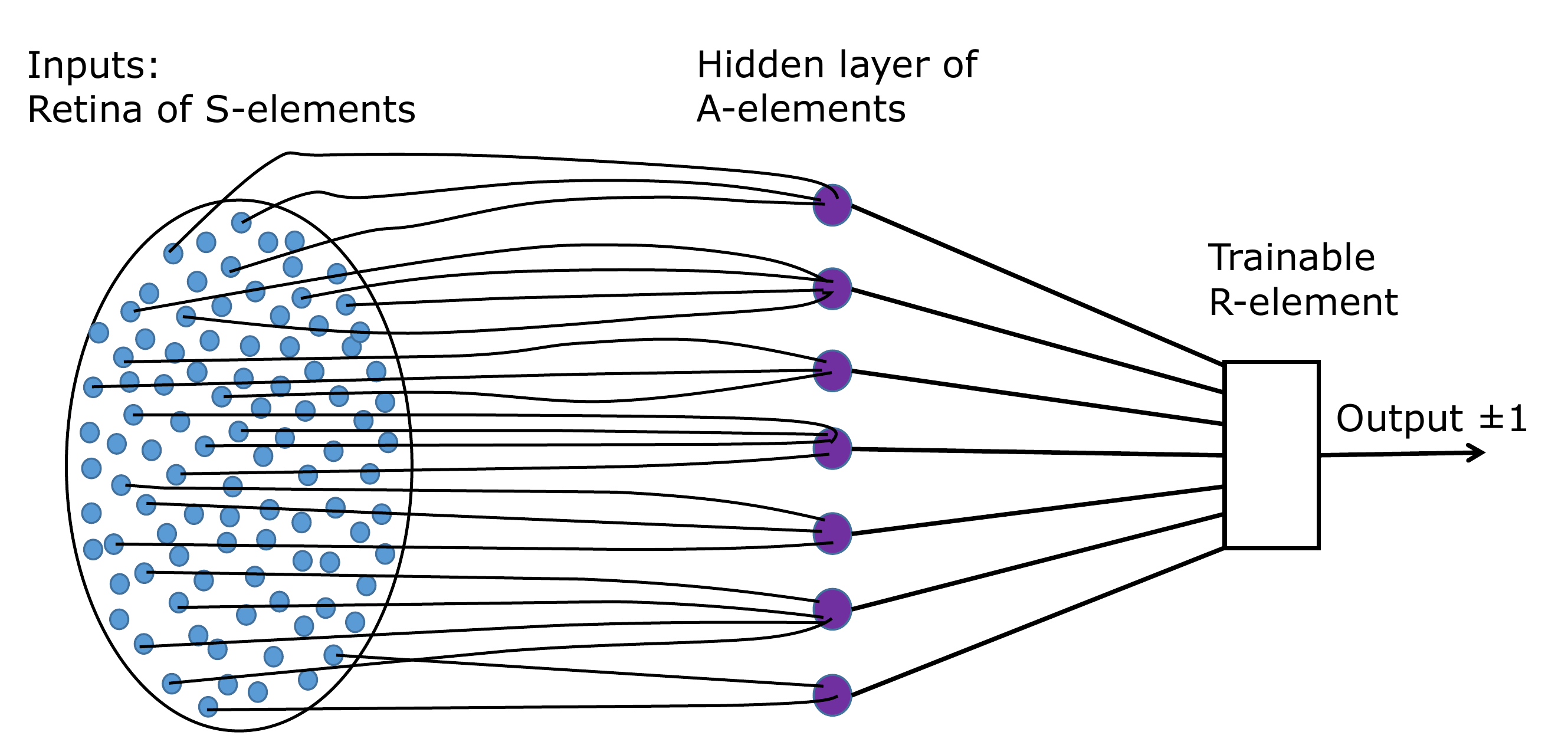}
\caption{Rosenblatt's elementary perceptron (re-drown from the Rosenblatt book \cite{Rosenblatt1962}). 
\label{Fig:ElementaryPerceptron}}
\end{figure}

Rosenblatt \cite{Rosenblatt1962} studied elementary perceptrons (Fig.~\ref{Fig:ElementaryPerceptron}). $A$- and $R$-elements are the  classical linear threshold neurons. The $R$ element is trainable by the Rosenblatt algorithm, while the $A$-elements should represent  a sufficient collection of features.

Rosenblatt assumed no restrictions on the choice of the $A$-elements. He proved that the elementary perceptrons can separate any two non-intersecting sets of binary images (Rosenblatt's Theorem 1 in  \cite{Rosenblatt1962}).  The proof was very simple. For each binary image $x$ we can create an $A$-element $A_x$ that produces output 1 for this image and 0 for all other. Indeed, let the input retina have $n$ elements and $x=(x_1,\ldots, x_n)$ be a binary vector ($x_i=0$ or 1) with $k$ non-zero elements. The corresponding $A$-element $A_x$ has input synapses with weights $w_i=1/k$ if $x_i=1$ and $w_i=-1/k$ if $x_i=0$. For an arbitrary binary image $y$, $$\sum w_i y_i \leq 1$$ and this sum is equal to $1$ if and only if $y=x$. The threshold for the $A_x$ output can be selected as $1-\frac{1}{2k}$. Thus,
\begin{equation}\label{Eq:Simplex}
OutA_x(y)=\left\{
\begin{array}{ll}
0, & if \;\sum w_i y_i <1-\frac{1}{2k};\\
1, & if \;\sum w_i y_i \geq 1-\frac{1}{2k}.
\end{array} \right.
\end{equation}

The set of neurons $A_x$ created for all binary vectors $x$ transforms binary images into the vertexes of the  standard simplex in $\mathbb{R}^{2^n}$ with coordinates $OutA_x(y)$ (\ref{Eq:Simplex}). Any two non-intersecting subsets of the standard simplex can be separated by a hyperplane. Therefore, there exists an $R$-element that separates them. According to the convergence theorem (Rosenblatt's Theorem 4 in  \cite{Rosenblatt1962}), this $R$-element can be found by the perceptron learning algorithm (a simple relaxation method for solving of systems of linear inequalities). 

Thus, first Rosenblatt's theorem is proven:

\begin{theorem}\label{Theorem:Ros1}
Elementary perceptron can separate any two non-intersecting sets of binary images.
\end{theorem}

Of course, selection of the $A$-elements in the proposed form {\it for all} $2^n$ binary images is not necessary in the realistic applied classification problems. Sometimes, even empty hidden layer can be used (the so-called linearly separable problems). Therefore, together with Rosenblatt's Theorem 1 we get a problem about reasonable (if not optimal) selection of $A$-element. There are many frameworks for approaching this question, for example the general feature selection algorithms: we generate (for example, randomly, or with some additional heuristics) large set of $A$-elements that is sufficient for solving classification problem, and then select the appropriate set of features using different methods. For the bibliography about feature selection we refer to recent reviews \cite{FeatureSel1, FeatureSel2, FeatureSel3}.

The Minsky and Papert book ``Perceptron'' \cite{Minsky1988} was published seven years later than Rosenblatt's book. They started from the restricted percetrons and assumed that each $A$-element has a bounded receptive field (either by a pre-selected diameter or by the bounded number of inputs). Immediately, instead of Rosenblatt's omnipotence of unrestricted perceptrons, they found that the elementary perceptron with such restrictions cannot solve some problems like connectivity of the image or parity of the number of pixels in it.  Minsky and Papert results were generalized to more general metric spaces and graphs \cite{Seifert1992}. The heuristic behind these results is quite simple: if a human cannot solve the problem immediately, by a glance, and needs to apply some sequential operations like counting pixels or following tangled path then this problem is not solvable by a restricted elementary perceptron. 

 At the same time, we can expect that the unrestricted elementary perceptron can solve this problem but for the cost of great (exponential?) complexity.
Multilayer (``deep'') networks are expected to solve these problems without explosion of complexity. In that sense, deep networks should be simpler than shallow networks for the problems that cannot be solved by restricted elementary perceptrons and require (from humans) combination of parallel and sequential actions.

\begin{figure}[h]\centering
\includegraphics[width=0.47\textwidth]{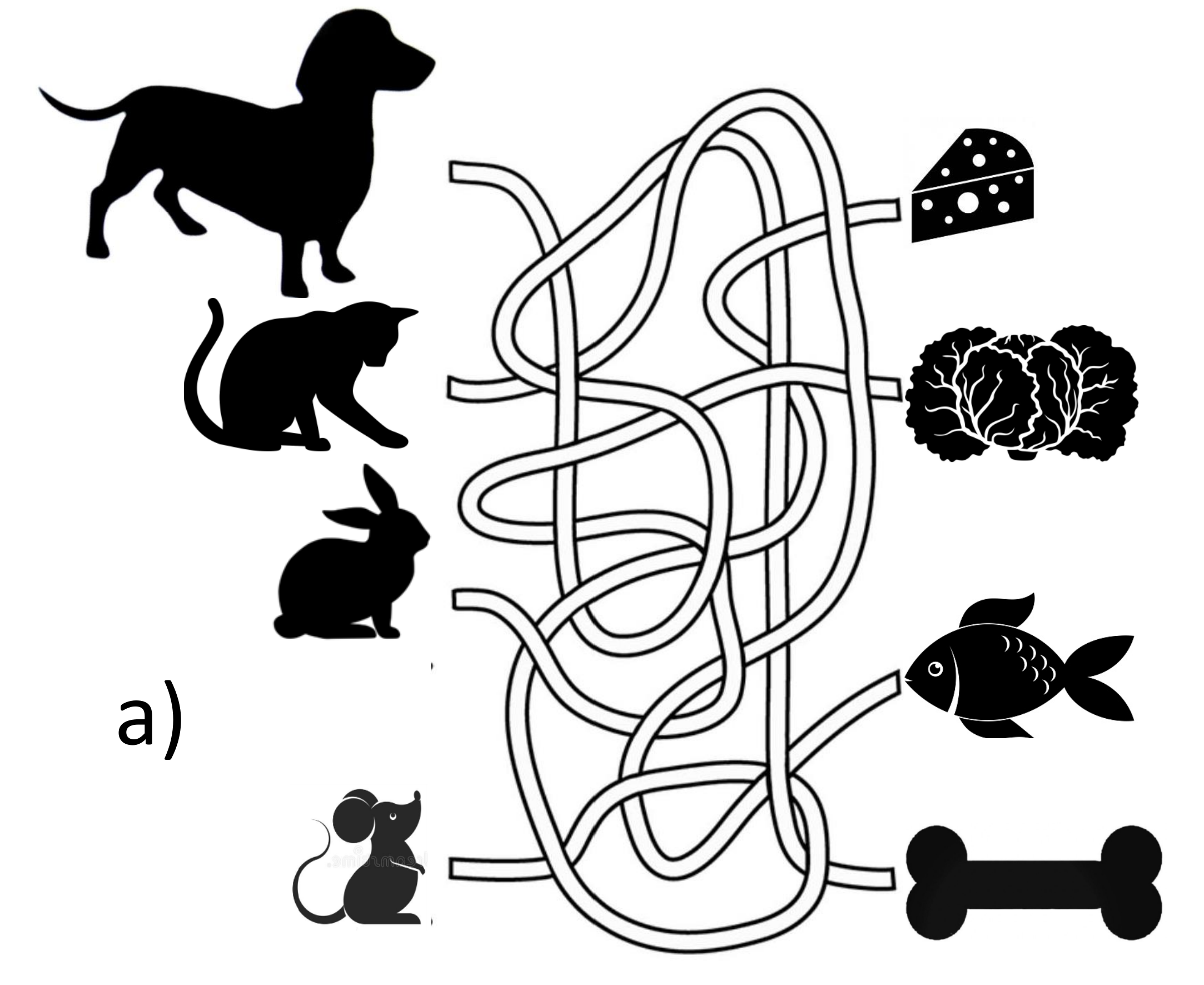}
\;\;
\includegraphics[width=0.47\textwidth]{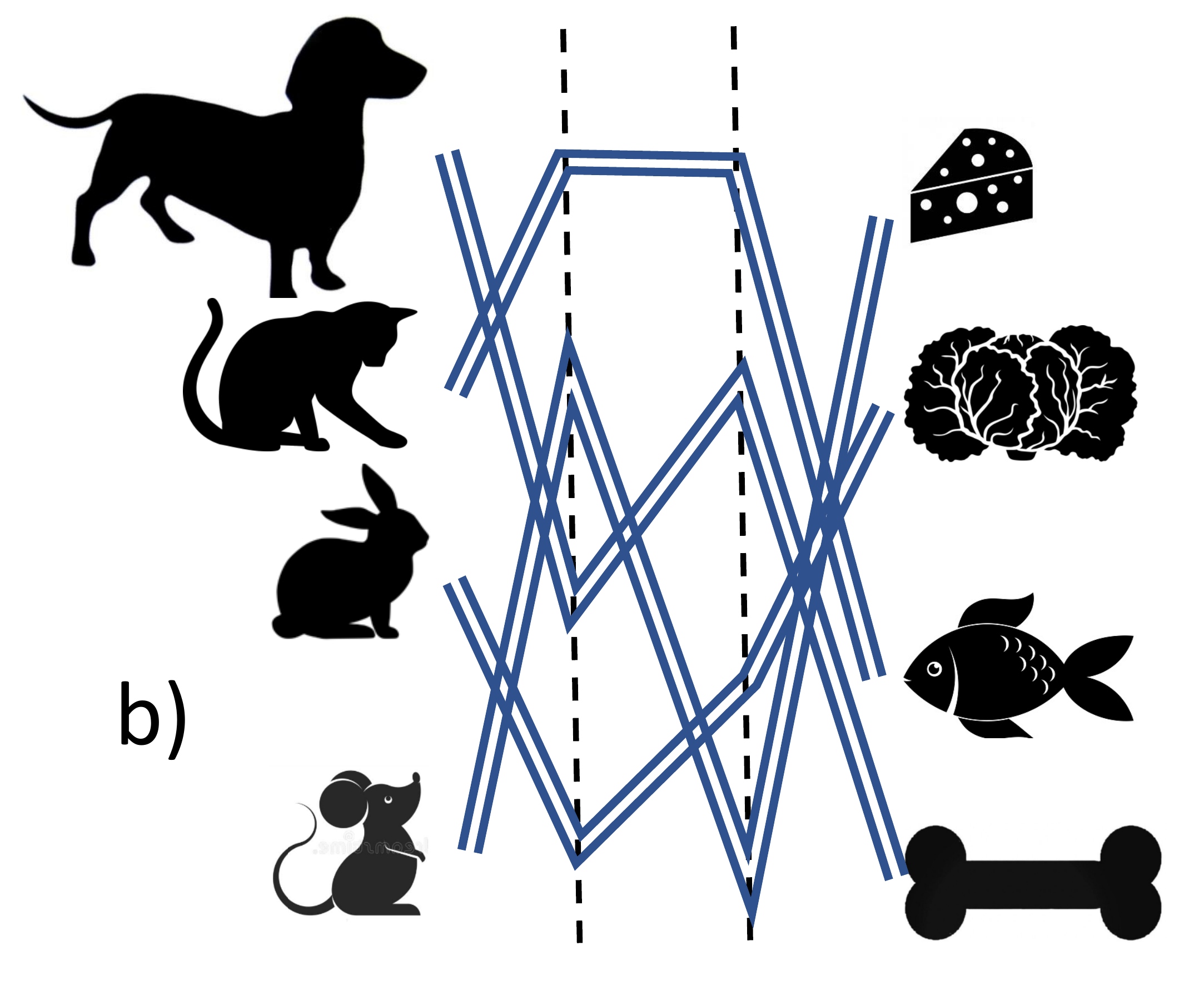}
\caption{Have we chosen the right delicacies (right) for our guests (left)? a)~A prototype travel maze problem.  b)~A simplified form of the problem with piece-wise linear paths for further formal description (Sec.~\ref{Sec:Formal}). Complexity depends on the number of guests and the number of links in a path.   
\label{Fig:MazeTravel}}
\end{figure}

In this note, we demonstrate the relative simplicity of deep solvers on a version of the  well-known travel maze problem (Fig.~\ref{Fig:MazeTravel}). This geometric problem is closely related to the connectivity problem and has been used for benchmarking in various areas of machine learning (see, for example, \cite{ElPiGraph2020}).

For formal analysis of the travel maze problem, we need to represent the paths on a discrete retina of $S$-elements (Fig.~\ref{Fig:ElementaryPerceptron}). Then, to implement the logic of the proof of Rosenblatt's first theorem, each $A$-element should be an indicator element for a possible path. For each guest-delicacy pair in Fig.~\ref{Fig:MazeTravel} a) or b), an elementary perceptron must be created that returns 1 if there is a path from this guest to this delicacy, and 0 if there are no such paths. Thus, $n^2$ elementary perceptrons should be created. We can easily combine them in a shallow network with  $n^2$ outputs. To finalize the formal statement we should specify the set of the possible paths. In our work, we select a very simple specification without loops, steps back or non-transversal intersections of paths (Fig.~\ref{Fig:MazeTravel} b)).

\section{Formal problem statement \label{Sec:Formal}}

\begin{center}
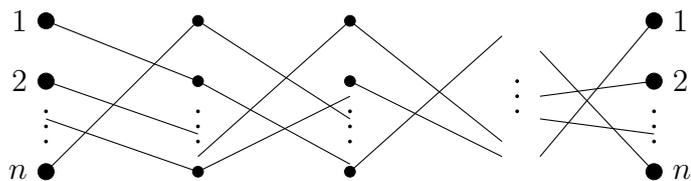
\begin{figure}[ht]
\centering  
\begin{tikzpicture}

\draw (-0.1,0.0) node[anchor=east] {$n$};
\draw (-0.1,1.2) node[anchor=east] {$2$};
\draw (-0.1,2.0) node[anchor=east] {$1$};

\filldraw  (0,0) circle (3.0pt);

\filldraw  (0.0,0.4) circle (0.5pt);
\filldraw  (0.0,0.6) circle (0.5pt);
\filldraw  (0.0,0.8) circle (0.5pt);

\filldraw  (0,1.2) circle (3.0pt);
\filldraw  (0,2.0) circle (3.0pt);

\draw (0,0) -- (2,2);
\draw (0,1.2) -- (2,0.5);
\draw (0,2.0) -- (2,1.2);
\draw (0,0.7) -- (2,0);


\filldraw  (2,0) circle (2.0pt);

\filldraw  (2,0.4) circle (0.5pt);
\filldraw  (2,0.6) circle (0.5pt);
\filldraw  (2,0.8) circle (0.5pt);

\filldraw  (2,1.2) circle (2.0pt);
\filldraw  (2,2.0) circle (2.0pt);

\draw (2,0) -- (4,1.0);
\draw (2,1.2) -- (4,0.1);
\draw (2,0.2) -- (4,2.0);
\draw (2,2) -- (4,0.7);


\filldraw  (4,0) circle (2.0pt);

\filldraw  (4,0.4) circle (0.5pt);
\filldraw  (4,0.6) circle (0.5pt);
\filldraw  (4,0.8) circle (0.5pt);

\filldraw  (4,1.2) circle (2.0pt);
\filldraw  (4,2.0) circle (2.0pt);

\draw (4,0) -- (6,1.8);
\draw (4,1.2) -- (6,0.2);
\draw (4,2.0) -- (6,0.4);


\filldraw  (6.2,0.8) circle (0.5pt);
\filldraw  (6.2,1.0) circle (0.5pt);
\filldraw  (6.2,1.2) circle (0.5pt);


\filldraw  (8,0) circle (3.0pt);

\filldraw  (8,0.4) circle (0.5pt);
\filldraw  (8,0.6) circle (0.5pt);
\filldraw  (8,0.8) circle (0.5pt);

\filldraw  (8,1.2) circle (3.0pt);
\filldraw  (8,2.0) circle (3.0pt);

\draw (6.5,1.6) -- (8,0);
\draw (6.5,1.0) -- (8,1.2);
\draw (6.5,0.2) -- (8,2.0);
\draw (6.5,0.7) -- (8,0.5);

\draw (8.1,0.0) node[anchor=west] {$n$};
\draw (8.1,1.2) node[anchor=west] {$2$};
\draw (8.1,2.0) node[anchor=west] {$1$};

\end{tikzpicture}
\caption{Game diagram with $ L $ stages (a formalized and simplified version of the travel maze problem).}   
\label{game}
\end{figure}
\end{center}

Consider the following problem. There are $n$ people, each of whom owns a single object from the set $\{1,2,\ldots,n\},$ and different people that own different objects. This correspondence between people and objects can be drawn as a diagram consisting of $n$ broken lines, each of which contains $L$ links (Fig. \ref{game}). Each stage of the diagram consists of $n$ links and can be encoded by a permutation or, equivalently, by permutation matrix in a natural way. Namely, if an edge is drawn from node $i$ to node $\pi(i)$ $(i=1,2, \ldots,n),$ then  the permutation can be represented in the form $$\pi=
\left(\begin{matrix} 
    1       & 2 & \cdots & n \\
    \pi(1)  & \pi(2) & \cdots & \pi(n)
\end{matrix}
\right)
$$ or by the permutation matrix $P=(p_{i,j}),$ where 
\begin{eqnarray*}
p_{i,j} = 
 \begin{cases}
   1, &\text{if}\ j=\pi(i)\\
   0, &\text{if}\  j\neq \pi(i).
 \end{cases} 
\end{eqnarray*}
If the permutation matrix $X_i$ $(i = 1,2, \ldots, n) $ corresponds to the $i$-th stage  then the product $ X_1 \cdot X_2 \cdot \ldots \cdot X_L $ is the permutation matrix again and it defines the correspondence ``person--object''. It is required to construct a shallow (fully connected) neural network that determines the correspondence ``person--object'' from the diagram. 

\section{Shallow neural network solution}

\begin{figure}[ht]
\centering  
\begin{tikzpicture}

\draw (0.1,7.3) node[anchor=west] {$Input\ layer$};
\draw (0.1,6.8) node[anchor=west] {$Ln^2\ neurons$};
\draw (3.6,7.3) node[anchor=west] {$Inner\ layer$};
\draw (3.8,6.8) node[anchor=west] {$(n!)^L\ neurons$};
\draw (7.8,7.3) node[anchor=west] {$Output\ layer$};
\draw (7.9,6.8) node[anchor=west] {$n^2\ neurons$};

\draw (0,0) -- (2.4,0);
\draw (0,0.8) -- (2.4,0.8);
\draw (0,1.6) -- (2.4,1.6);
\draw (0,2.4) -- (2.4,2.4);

\draw (0,0) -- (0,2.4);
\draw (0.8,0) -- (0.8,2.4);
\draw (1.6,0) -- (1.6,2.4);
\draw (2.4,0) -- (2.4,2.4);

\draw (-0.10,2.0) node[anchor=west] {$x_{1,1}^{(L)}$};
\draw (0.8,2.0) node[anchor=west] {$\cdots$};
\draw (1.50,2.0) node[anchor=west] {$x_{1,n}^{(L)}$};
\draw (0.2,1.3) node[anchor=west] {$\vdots$};
\draw (0.8,1.3) node[anchor=west] {$\ddots$};
\draw (1.8,1.3) node[anchor=west] {$\vdots$};
\draw (-0.10,0.4) node[anchor=west] {$x_{n,1}^{(L)}$};
\draw (0.8,0.4) node[anchor=west] {$\cdots$};
\draw (1.50,0.4) node[anchor=west] {$x_{n,n}^{(L)}$};

\filldraw  (1.2,2.8) circle (1pt);
\filldraw  (1.2,3.2) circle (1pt);
\filldraw  (1.2,3.6) circle (1pt);

\draw[->] (2.4,1.2) -- (4.5,5.5); 
\draw[->] (2.4,1.2) -- (4.5,3.8); 
\draw[->] (2.4,1.2) -- (4.5,0.8); 

\draw[->] (2.4,5.2) -- (4.5,5.7); 
\draw[->] (2.4,5.2) -- (4.5,3.9); 
\draw[->] (2.4,5.2) -- (4.6,1.2); 

\draw[->] (5.9,0.8) -- (8.0,2.9); 
\draw[->] (5.9,3.8) -- (8.0,3.0); 
\draw[->] (5.9,5.5) -- (8.0,3.1); 

\draw (0,4) -- (2.4,4);
\draw (0,4.8) -- (2.4,4.8);
\draw (0,5.6) -- (2.4,5.6);
\draw (0,6.4) -- (2.4,6.4);

\draw (0,4) -- (0,6.4);
\draw (0.8,4) -- (0.8,6.4);
\draw (1.6,4) -- (1.6,6.4);
\draw (2.4,4) -- (2.4,6.4);

\draw (-0.10,6.0) node[anchor=west] {$x_{1,1}^{(1)}$};
\draw (0.8,6.0) node[anchor=west] {$\cdots$};
\draw (1.50,6.0) node[anchor=west] {$x_{1,n}^{(1)}$};
\draw (0.2,5.3) node[anchor=west] {$\vdots$};
\draw (0.8,5.3) node[anchor=west] {$\ddots$};
\draw (1.8,5.3) node[anchor=west] {$\vdots$};
\draw (-0.10,4.4) node[anchor=west] {$x_{n,1}^{(1)}$};
\draw (0.8,4.4) node[anchor=west] {$\cdots$};
\draw (1.50,4.4) node[anchor=west] {$x_{n,n}^{(1)}$};

\draw  (5.2,3.8) circle (20pt);
\draw  (5.2,5.7) circle (20pt);
\filldraw  (5.2,1.8) circle (1pt);
\filldraw  (5.2,2.2) circle (1pt);
\filldraw  (5.2,2.6) circle (1pt);
\draw  (5.2,0.8) circle (20pt);

\draw (4.8,3.8) node[anchor=west] {$y_1$};
\draw (4.8,5.7) node[anchor=west] {$y_0$};
\draw (4.4,0.8) node[anchor=west] {$y_{(n!)^L-1}$};

\draw (8,2) -- (10.4,2);
\draw (8,2.8) -- (10.4,2.8);
\draw (8,3.6) -- (10.4,3.6);
\draw (8,4.4) -- (10.4,4.4);

\draw (8,2) -- (8,4.4);
\draw (8.8,2) -- (8.8,4.4);
\draw (9.6,2) -- (9.6,4.4);
\draw (10.4,2) -- (10.4,4.4);

\draw (7.9,4.0) node[anchor=west] {$z_{1,1}$};
\draw (8.8,4.0) node[anchor=west] {$\cdots$};
\draw (9.5,4.0) node[anchor=west] {$z_{1,n}$};
\draw (8.2,3.3) node[anchor=west] {$\vdots$};
\draw (8.8,3.3) node[anchor=west] {$\ddots$};
\draw (9.8,3.3) node[anchor=west] {$\vdots$};
\draw (7.9,2.4) node[anchor=west] {$z_{n,1}$};
\draw (8.8,2.4) node[anchor=west] {$\cdots$};
\draw (9.5,2.4) node[anchor=west] {$z_{n,n}$};

\end{tikzpicture}
\caption{A shallow (fully connected) neural network for the travel maze problem 
(Fig.~\ref{game}). It differs from the classical elementary perceptron 
(Fig.~\ref{Fig:ElementaryPerceptron}) by $n^2$ output neurons instead of one, and can be considered as a union of $n^2$ elementary perceptrons with joint retina and hidden layer of $A$-elements.}   
\label{neuro_net}
\end{figure}
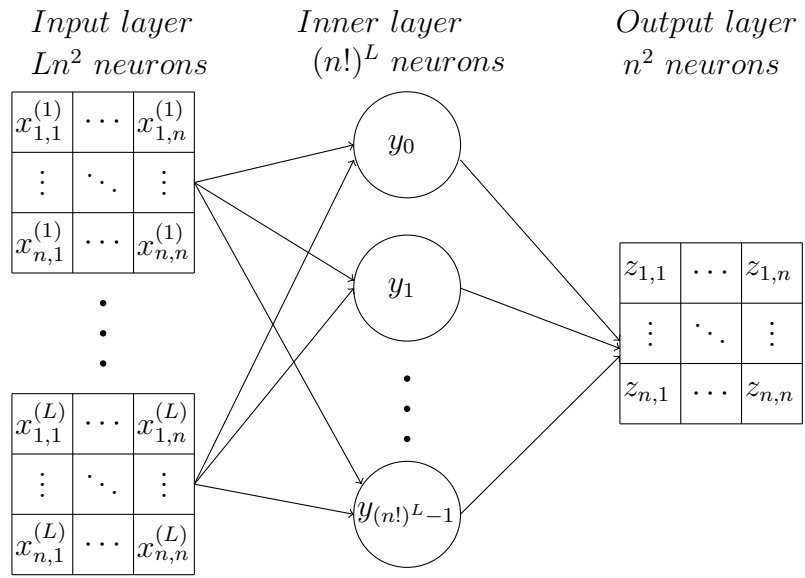

Arrange all the $ n! $ permutations $S_n=\{\pi_0, \pi_1, \ldots, \pi_{n!-1}\}.$ Then $ M = \{P_0, P_1, \ldots, P_{n!-1} \} $ is the set of the corresponding $ n \times n$ permutation matrices. 

We denote the entries of the matrix $P_k$ by $p_{i, j}^{(k)},$ 
where $$ p_{1, \pi_k (1)}^{(k)} = p_{2, \pi_k(2)}^{(k)} = \ldots = p_{n, \pi_k(n)}^{(k)} = 1, $$ and the other entries are equal to $0.$

Let an $L$-tuple $(X_1,X_2,\ldots,X_L)$ ($X_i\in M$ for all $i=1,\ldots, L$) be  an $L$-tuple (a word  of length $L$)  over the set of  permutation matrices $M$. The number of such permutation matrices is $n!$, and the number of $L$-tuples with elements from $M$ is $(n!)^{L} $. Consider  all such  words arranged by the lexicographical order. For each word $W_j=(X_1,X_2,\ldots,X_L)$ with the number $j$ $(j=0,1, \ldots,(n!)^{L}-1)$ we assign the same number to the product $X_1\cdot X_2\cdot\ldots\cdot X_L =P_{t_j}$ The matrix $P_{t_j}$ is also a $n\times n$ permutation matrix. 

Entries of matrices $X_1,X_2,\ldots,X_L$ are inputs of the neural network (see Fig. \ref{neuro_net}). Each input corresponds to an input neuron ($S$-element, Fig.~\ref{Fig:ElementaryPerceptron}).  An inner layer $A$-neuron $y_j$ corresponds to the $L$-tuple $W_j=(X_1,X_2,\ldots,X_L)$ having the same number  $j$. The neuron $y_j$ should give the output signal 1, if the input vector is $W_j$ and output 0 for all other $(n!)^{L} -1$ possible input vectors. Other input vectors are impossible in our settings (Fig. \ref{neuro_net}). Each matrix element of every permutation matrix $X_i$ is either 0 or 1, therefore the $L$-tuples star of output connections of the inner neuron can be coded as a $0-1$ sequence, that is a vertex of the $Ln^2$-dimensional unit cube. (Apparently, there are more vertices than $L$-tuples of permutation matrices.) This cube is a convex body and each vertex can be separated from all other vertices by a linear functional. In particular, for each $j$ we can find such a linear functional $l_j$ that $l_j(W_j)>1/2$ and $l_j(W_k)<1/2$ ($k\neq j$, $k,j=1,\ldots, (n!)^{L}$). Here we, with some abuse of language, use the same notation for the tuple $W_j$ and the correspondent vertex of the cube (a $0-1$ sequence of the length $Ln^2$). Thus, each inner neuron $y_j$ can be chosen in the form of the linear threshold element with the output signal (compare to (\ref{Eq:Simplex} and Theorem~\ref{Theorem:Ros1}): 
$$y_j(W)=h(l_j(W)-1/2),$$
where $h$ is the Heaviside step function.
We use for the output of the neuron $y_j$ the same notation $y_j$. 

The structural difference of the shallow network (Fig.~\ref{neuro_net}) for the travel maze problem from the elementary perceptron (Fig.~\ref{Fig:ElementaryPerceptron}) is the number of neurons in the output layer. For the travel maze problem the answer is the permutation matrix with $n^2$ $0-1$ elements. The inner layer neuron $y_j$ detects the $L$-tuple of one-step permutation matrices $W_j=(X_1,X_2,\ldots,X_L)$. When this input vector is detected, $y_i$ sends the output signal 1 to the output neurons connected with it.   For all other input vectors, it keeps silent. The output neurons are just simple linear adders. The output neurons $z_{qr}$ are labelled by pairs of indexes, $q,r=1,\ldots , n$. The matrix of outputs is the permutation matrix from the start to the end of the travel. The structure of the output connections of  $y_i$ is determined by the input $L$-tuple $W_j=(X_1,X_2,\ldots,X_L)$: the connection from $y_j$ to $z_{qr}$ has weight 1, if the corresponding entry $(P_{t_j})_{qr}=1$ and is 0 if $(P_{t_j})_{qr}=0$. (Recall that $P_{t_j}=X_1\cdot X_2\cdot\ldots\cdot X_L $.)

Thus the neuron $y_j$ corresponds to our problem answer. Let us represent the network functioning in more detail with explicit algebraic presentations. All the inputs and outputs are Boolean ($0-1$) variables. We use the standard Boolean algebra notations. In particular, $\overline{x}=1-x$ 
 
\begin{eqnarray*}
y_0&\longrightarrow& \overbrace{P_0\cdot \ldots\cdot P_0\cdot P_0}^{L}=P_{t_0},\\
y_1&\longrightarrow& P_0\cdot \ldots\cdot P_0\cdot P_1=P_{t_1},\\&\ldots&\\
y_{n!-1}&\longrightarrow& P_0\cdot \ldots\cdot P_0\cdot P_{n!-1}=P_{t_{n!-1}},\\
y_{n!}&\longrightarrow& P_0\cdot \ldots\cdot P_1\cdot P_{0}=P_{t_{n!}},\\&\ldots&\\
y_{2n!-1}&\longrightarrow& P_0\cdot \ldots\cdot P_1\cdot P_{n!-1}=P_{t_{2n!-1}},\\&\ldots&\\
y_{(n!)^L-1}&\longrightarrow& P_{n!-1}\cdot \ldots\cdot P_{n!-1}\cdot P_{n!-1}=P_{t_{(n!)^L-1}}
\end{eqnarray*}
Thus, if $y_j$ is a neuron of the inner layer, then it corresponds to the product
$$P_{t_j}=P_{a_{j,L-1}}\cdot P_{a_{j,L-2}}\cdot\ldots\cdot P_{a_{j,1}}\cdot P_{a_{j,0}},$$ where $$j=a_{j,L-1}(n!)^{L-1}+a_{j,L-2}(n!)^{L-2}+\ldots +a_{j,1}(n!)+a_{j,0}$$ --- expansion of $j$ in the base $n!.$ 

We need $$y_j(X_1, X_2,\ldots, X_L)=1\ \ \Longleftrightarrow\ \ (X_1,X_2,\ldots, X_L)=(P_{a_{j,L-1}}, P_{a_{j,L-2}}, \ldots, P_{a_{j,0}}).$$
Denote $$I_{k}=\{j:~ t_j=k\},\ (k=0,\ldots,n!-1),$$ 
$$M_{ij}=\{k:~ \pi_k(i)=j\}=\{k:~ p_{i,j}^{(k)}=1\}.$$ Note that $|M_{ij}|=(n-1)!$.
 
 Each neuron $y_j$ of the inner layer for $j\in I_{k}$ corresponds to the same product $P_k$: 
  
$$y_j=x_{1,\pi_{a_{j,L-1}}(1)}^{(1)}\cdot x_{2,\pi_{a_{j,L-1}}(2)}^{(1)}\ldots x_{n,\pi_{a_{j,L-1}}(n)}^{(1)}
\cdot x_{1,\pi_{a_{j,L-2}}(1)}^{(2)}\cdot x_{2,\pi_{a_{j,L-2}}(2)}^{(2)}\ldots x_{n,\pi_{a_{j,L-2}}(n)}^{(2)}\cdot\ldots$$
$$\cdot x_{1,\pi_{a_{j,0}}(1)}^{(L)}\cdot x_{2,\pi_{a_{j,0}}(2)}^{(L)}\ldots x_{n,\pi_{a_{j,0}}(n)}^{(L)}\cdot\prod\limits_{(\alpha,\beta)\neq(i,\pi_{a_{j,s}}(i))}\overline{x_{\alpha,\beta}^{(\gamma)}}.$$ 
The third level neurons  $z_{ij}$  form the matrix $Z=(z_{ij})$ that is the answer to this problem:

\begin{eqnarray}\label{zij}
z_{i,j}=\bigvee\limits_{s\in \bigcup\limits_{k\in M_{ij}}I_{k}}y_s,\ (i,j=1,\ldots,n).
\end{eqnarray}
Since $|M_{ij}| = (n-1)!,$ $|I_k| = (n!)^{L-1}, $ then the right-hand side in the equality $(\ref{zij})$ contains exactly $(n-1)!\cdot (n!)^{L-1} $ terms $y_s.$
\begin{theorem}\label{ShallowEstimate}
The constructed shallow neural network has a depth of $3$,
$$(L + 1)n^2 + (n!)^L $$ neurons, and $$(L + 1) n^2(n!)^L$$ connections between neurons.
\end{theorem}

The constructed network memorizes products in all $L$-tuples of permutation matrices, recognizes the input $L$-tuple of permutations, and sends the product to the output.

\begin{example} Consider $n=2, L=3.$ Then $$\pi_0=\begin{pmatrix} 
    1  & 2 \\
    1  & 2
\end{pmatrix}, \pi_1=\begin{pmatrix} 
    1  & 2 \\
    2  & 1
\end{pmatrix}, 
P_0=\begin{pmatrix}
    1  & 0 \\
    0  & 1
\end{pmatrix}, P_1=\begin{pmatrix}
    0  & 1 \\
    1  & 0
\end{pmatrix}.$$

\begin{center}
\begin{figure}[ht]
\centering  
\begin{tikzpicture}
	\filldraw  (0,0) circle (2.0pt);
	\filldraw  (0,1) circle (2.0pt);
	
	\draw (0,0) -- (1,0);
	\draw (0,1) -- (1,1);
	
	\filldraw  (1,0) circle (1pt);
	\filldraw  (1,1) circle (1pt);
	
	\draw (1,0) -- (2,0);
	\draw (1,1) -- (2,1);
	
	\filldraw  (2,0) circle (1pt);
	\filldraw  (2,1) circle (1pt);
	
	\draw (2,0) -- (3,0);
	\draw (2,1) -- (3,1);
	
	\filldraw  (3,0) circle (2pt);
	\filldraw  (3,1) circle (2pt);
	
	\filldraw  (4,0) circle (2.0pt);
	\filldraw  (4,1) circle (2.0pt);
	
	\draw (4,0) -- (5,0);
	\draw (4,1) -- (5,1);
	
	\filldraw  (5,0) circle (1pt);
	\filldraw  (5,1) circle (1pt);
	
	\draw (5,0) -- (6,0);
	\draw (5,1) -- (6,1);
	
	\filldraw  (6,0) circle (1pt);
	\filldraw  (6,1) circle (1pt);
	
	\draw (6,0) -- (7,1);
	\draw (6,1) -- (7,0);
	
	\filldraw  (7,0) circle (2pt);
	\filldraw  (7,1) circle (2pt);
	
	\filldraw  (8,0) circle (2.0pt);
	\filldraw  (8,1) circle (2.0pt);
	
	\draw (8,0) -- (9,0);
	\draw (8,1) -- (9,1);
	
	\filldraw  (9,0) circle (1pt);
	\filldraw  (9,1) circle (1pt);
	
	\draw (9,0) -- (10,1);
	\draw (9,1) -- (10,0);
	
	\filldraw  (10,0) circle (1pt);
	\filldraw  (10,1) circle (1pt);
	
	\draw (10,0) -- (11,0);
	\draw (10,1) -- (11,1);
	
	\filldraw  (11,0) circle (2pt);
	\filldraw  (11,1) circle (2pt);
	
	\filldraw  (12,0) circle (2.0pt);
	\filldraw  (12,1) circle (2.0pt);
	
	\draw (12,0) -- (13,0);
	\draw (12,1) -- (13,1);
	
	\filldraw  (13,0) circle (1pt);
	\filldraw  (13,1) circle (1pt);
	
	\draw (13,0) -- (14,1);
	\draw (13,1) -- (14,0);
	
	\filldraw  (14,0) circle (1pt);
	\filldraw  (14,1) circle (1pt);
	
	\draw (14,0) -- (15,1);
	\draw (14,1) -- (15,0);
	
	\filldraw  (15,0) circle (2pt);
	\filldraw  (15,1) circle (2pt);
	
	\filldraw  (0,-2) circle (2.0pt);
	\filldraw  (0,-1) circle (2.0pt);
	
	\draw (0,-2) -- (1,-1);
	\draw (0,-1) -- (1,-2);
	
	\filldraw  (1,-2) circle (1pt);
	\filldraw  (1,-1) circle (1pt);
	
	\draw (1,-2) -- (2,-2);
	\draw (1,-1) -- (2,-1);
	
	\filldraw  (2,-2) circle (1pt);
	\filldraw  (2,-1) circle (1pt);
	
	\draw (2,-2) -- (3,-2);
	\draw (2,-1) -- (3,-1);
	
	\filldraw  (3,-2) circle (2pt);
	\filldraw  (3,-1) circle (2pt);
	
	\filldraw  (4,-2) circle (2.0pt);
	\filldraw  (4,-1) circle (2.0pt);
	
	\draw (4,-2) -- (5,-1);
	\draw (4,-1) -- (5,-2);
	
	\filldraw  (5,-2) circle (1pt);
	\filldraw  (5,-1) circle (1pt);
	
	\draw (5,-2) -- (6,-2);
	\draw (5,-1) -- (6,-1);
	
	\filldraw  (6,-2) circle (1pt);
	\filldraw  (6,-1) circle (1pt);
	
	\draw (6,-2) -- (7,-1);
	\draw (6,-1) -- (7,-2);
	
	\filldraw  (7,-2) circle (2pt);
	\filldraw  (7,-1) circle (2pt);
	
	\filldraw  (8,-2) circle (2.0pt);
	\filldraw  (8,-1) circle (2.0pt);
	
	\draw (8,-2) -- (9,-1);
	\draw (8,-1) -- (9,-2);
	
	\filldraw  (9,-2) circle (1pt);
	\filldraw  (9,-1) circle (1pt);
	
	\draw (9,-2) -- (10,-1);
	\draw (9,-1) -- (10,-2);
	
	\filldraw  (10,-2) circle (1pt);
	\filldraw  (10,-1) circle (1pt);
	
	\draw (10,-2) -- (11,-2);
	\draw (10,-1) -- (11,-1);
	
	\filldraw  (11,-2) circle (2pt);
	\filldraw  (11,-1) circle (2pt);
	
	\filldraw  (12,-2) circle (2.0pt);
	\filldraw  (12,-1) circle (2.0pt);
	
	\draw (12,-2) -- (13,-1);
	\draw (12,-1) -- (13,-2);
	
	\filldraw  (13,-2) circle (1pt);
	\filldraw  (13,-1) circle (1pt);
	
	\draw (13,-2) -- (14,-1);
	\draw (13,-1) -- (14,-2);
	
	\filldraw  (14,-2) circle (1pt);
	\filldraw  (14,-1) circle (1pt);
	
	\draw (14,-2) -- (15,-1);
	\draw (14,-1) -- (15,-2);
	
	\filldraw  (15,-2) circle (2pt);
	\filldraw  (15,-1) circle (2pt);
	
	\draw  (1.5,0.5) circle (10pt);
	\draw (1.2,0.5) node[anchor=west] {$y_0$};
	\draw (0.5,1.5) node[anchor=west] {$P_{0}P_{0}P_{0}=P_{0}$};
	
	\draw  (1.5,-1.5) circle (10pt);
	\draw (1.2,-1.5) node[anchor=west] {$y_4$};
	\draw (4.5,1.5) node[anchor=west] {$P_{0}P_{0}P_{1}=P_{1}$};
	
	\draw  (8.5,0.5) circle (10pt);
	\draw (8.2,0.5) node[anchor=west] {$y_2$};
	\draw (8.5,1.5) node[anchor=west] {$P_{0}P_{1}P_{0}=P_{1}$};
	
	\draw  (12.5,0.5) circle (10pt);
	\draw (12.2,0.5) node[anchor=west] {$y_3$};
	\draw (12.5,1.5) node[anchor=west] {$P_{0}P_{1}P_{1}=P_{0}$};
	
	\draw  (5.2,0.5) circle (10pt);
	\draw (4.9,0.5) node[anchor=west] {$y_1$};
	\draw (0.5,-0.5) node[anchor=west] {$P_{1}P_{0}P_{0}=P_{1}$};
	
	\draw  (5.5,-1.5) circle (10pt);
	\draw (5.2,-1.5) node[anchor=west] {$y_5$};
	\draw (4.5,-0.5) node[anchor=west] {$P_{1}P_{0}P_{1}=P_{0}$};
	
	\draw  (10.5,-1.5) circle (10pt);
	\draw (10.2,-1.5) node[anchor=west] {$y_6$};
	\draw (8.5,-0.5) node[anchor=west] {$P_{1}P_{1}P_{0}=P_{0}$};
	
	\draw  (15.3,-1.5) circle (10pt);
	\draw (15.0,-1.5) node[anchor=west] {$y_7$};
	\draw (12.5,-0.5) node[anchor=west] {$P_{1}P_{1}P_{1}=P_{1}$};
	
\end{tikzpicture}
\caption{The case $n=2, L=3$.}   
\label{game_2}
\end{figure}
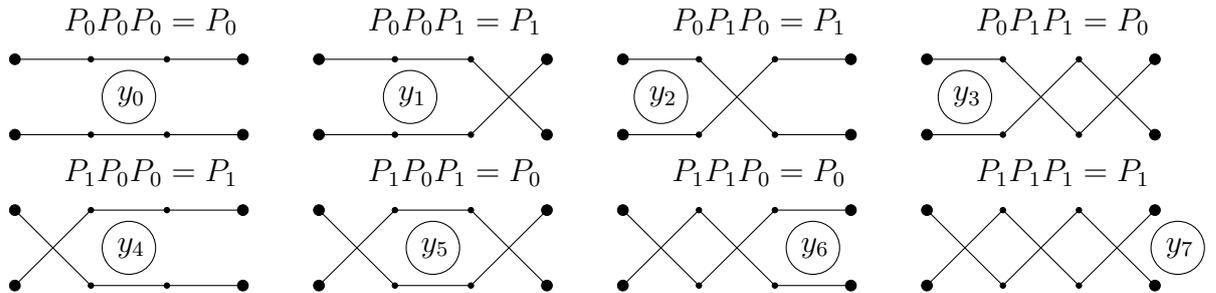
\end{center}

$$X_1=\begin{pmatrix}
    x_{11}^{(1)}  &  x_{12}^{(1)} \\
    x_{21}^{(1)}  &  x_{22}^{(1)}
\end{pmatrix}, X_2=\begin{pmatrix}
    x_{11}^{(2)}  &  x_{12}^{(2)} \\
    x_{21}^{(2)}  &  x_{22}^{(2)}
\end{pmatrix}, X_3=\begin{pmatrix}
    x_{11}^{(3)}  &  x_{12}^{(3)} \\
    x_{21}^{(3)}  &  x_{22}^{(3)}
\end{pmatrix}.$$
For example, we have  $5=1\cdot(2!)^2+0\cdot(2!)+1$ for $j=5$ therefore
$$y_5=x_{1,\pi_1(1)}^{(1)}\cdot x_{2,\pi_1(2)}^{(1)}\cdot x_{1,\pi_0(1)}^{(2)}\cdot x_{2,\pi_0(2)}^{(2)}\cdot x_{1,\pi_1(1)}^{(3)}\cdot x_{2,\pi_1(2)}^{(3)}\cdot\prod\limits_{(\alpha,\beta)\neq(i,\pi_{a_{j,s}}(i))}\overline{x_{\alpha,\beta}^{(\gamma)}}=$$
$$=x_{1,2}^{(1)}\cdot x_{2,1}^{(1)}\cdot x_{1,1}^{(2)}\cdot x_{2,2}^{(2)}\cdot x_{1,2}^{(3)}\cdot x_{2,1}^{(3)}\cdot
\overline{x_{1,1}^{(1)}}\cdot\overline{x_{2,2}^{(1)}}\cdot
\overline{x_{1,2}^{(2)}}\cdot\overline{x_{2,1}^{(2)}}\cdot
\overline{x_{1,1}^{(3)}}\cdot\overline{x_{2,2}^{(3)}}.$$
We can write similar expressions for all other $y_j.$

In this case, we have $I_{0}=\{0,3,5,6\},$ $I_{1}=\{1,2,4,7\}$ and $M_{11}=\{0\}, M_{12}=\{1\}, M_{21}=\{1\},M_{22}=\{0\},$ so
$$z_{1,1}=z_{2,2}=y_0\vee y_3\vee y_5\vee y_6, $$
$$z_{1,2}=z_{2,1}=y_1\vee y_2\vee y_4\vee y_7.$$
\end{example}

\begin{example} Let $n=3, L=2.$ Then $$\pi_0=\begin{pmatrix}
    1  & 2 & 3 \\
    1  & 2 & 3
\end{pmatrix}, \pi_1=\begin{pmatrix}
    1  & 2 & 3\\
    1  & 3 & 2
\end{pmatrix}, 
\pi_2=\begin{pmatrix}
    1  & 2 & 3 \\
    2  & 1 & 3
\end{pmatrix},$$ $$\pi_3=\begin{pmatrix}
    1  & 2 & 3\\
    2  & 3 & 1
\end{pmatrix}, 
\pi_4=\begin{pmatrix}
    1  & 2 & 3 \\
    3  & 1 & 2
\end{pmatrix}, \pi_5=\begin{pmatrix}
    1  & 2 & 3\\
    3  & 2 & 1
\end{pmatrix}$$ 
$$P_0=\begin{pmatrix}
    1  & 0 & 0\\
    0  & 1 & 0\\
    0  & 0 & 1\\
\end{pmatrix}, P_1=\begin{pmatrix}
    1  & 0 & 0\\
    0  & 0 & 1\\
    0  & 1 & 0\\
\end{pmatrix}, P_2=\begin{pmatrix}
    0  & 1 & 0\\
    1  & 0 & 0\\
    0  & 0 & 1\\
\end{pmatrix},$$ $$P_3=\begin{pmatrix}
    0  & 1 & 0\\
    0  & 0 & 1\\
    1  & 0 & 0\\
\end{pmatrix}, P_4=\begin{pmatrix}
    0  & 0 & 1\\
    1  & 0 & 0\\
    0  & 1 & 0\\
\end{pmatrix}, P_5=\begin{pmatrix}
    0  & 0 & 1\\
    0  & 1 & 0\\
    1  & 0 & 0\\
\end{pmatrix}.$$

\begin{center}
\begin{tabular}{|c||c|c|c|c|c|c|}
	\hline
     $\cdot$&$P_0$&$P_1$&$P_2$&$P_3$&$P_4$&$P_5$  \\
     \hline
     \hline
     $P_0$&$P_0$&$P_1$&$P_2$&$P_3$&$P_4$&$P_5$\\
     \hline
     $P_1$&$P_1$&$P_0$&$P_3$&$P_2$&$P_5$&$P_4$\\
     \hline
     $P_2$&$P_2$&$P_4$&$P_0$&$P_5$&$P_1$&$P_3$\\
     \hline
     $P_3$&$P_3$&$P_5$&$P_1$&$P_4$&$P_0$&$P_2$\\
     \hline
     $P_4$&$P_4$&$P_2$&$P_5$&$P_0$&$P_3$&$P_1$\\
     \hline
     $P_5$&$P_5$&$P_3$&$P_4$&$P_1$&$P_2$&$P_0$\\
     \hline
\end{tabular}
\end{center}

\begin{center}
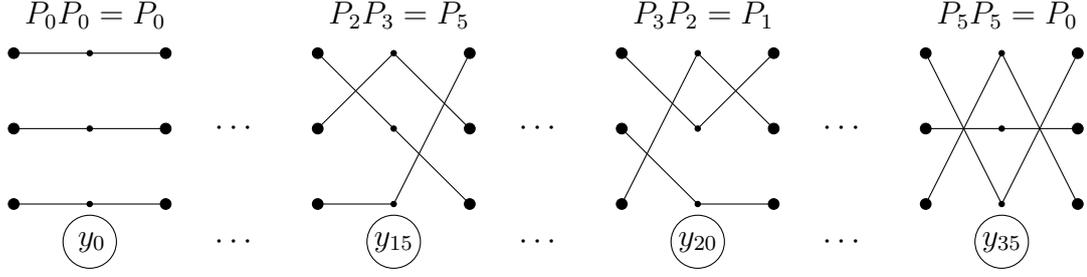
\begin{figure}[ht]
\centering  
\begin{tikzpicture}
\draw (0,2.5) node[anchor=west] {$P_{0}P_{0}=P_{0}$};
\filldraw  (0,0) circle (2.0pt);
\filldraw  (0,1) circle (2.0pt);
\filldraw  (0,2) circle (2.0pt);

\draw (0,0) -- (1,0);
\draw (0,1) -- (1,1);
\draw (0,2) -- (1,2);

\filldraw  (1,0) circle (1pt);
\filldraw  (1,1) circle (1pt);
\filldraw  (1,2) circle (1pt);

\draw (1,0) -- (2,0);
\draw (1,1) -- (2,1);
\draw (1,2) -- (2,2);

\filldraw  (2,0) circle (2pt);
\filldraw  (2,1) circle (2pt);
\filldraw  (2,2) circle (2pt);

\draw  (1,-0.5) circle (10pt);
\draw (0.7,-0.5) node[anchor=west] {$y_0$};

\draw (2.5,1) node[anchor=west] {$\cdots$};
\draw (2.5,-0.5) node[anchor=west] {$\cdots$};
\draw (4,2.5) node[anchor=west] {$P_{2}P_{3}=P_{5}$};
\filldraw  (4,0) circle (2.0pt);
\filldraw  (4,1) circle (2.0pt);
\filldraw  (4,2) circle (2.0pt);

\draw (4,0) -- (5,0);
\draw (4,1) -- (5,2);
\draw (4,2) -- (5,1);

\filldraw  (5,0) circle (1pt);
\filldraw  (5,1) circle (1pt);
\filldraw  (5,2) circle (1pt);

\draw (5,0) -- (6,2);
\draw (5,1) -- (6,0);
\draw (5,2) -- (6,1);

\filldraw  (6,0) circle (2pt);
\filldraw  (6,1) circle (2pt);
\filldraw  (6,2) circle (2pt);

\draw (6.5,1) node[anchor=west] {$\cdots$};

\draw  (5,-0.5) circle (10pt);
\draw (4.6,-0.5) node[anchor=west] {$y_{15}$};
\draw (6.5,-0.5) node[anchor=west] {$\cdots$};
\draw (8,2.5) node[anchor=west] {$P_{3}P_{2}=P_{1}$};
\filldraw  (8,0) circle (2.0pt);
\filldraw  (8,1) circle (2.0pt);
\filldraw  (8,2) circle (2.0pt);

\draw (8,0) -- (9,2);
\draw (8,1) -- (9,0);
\draw (8,2) -- (9,1);

\filldraw  (9,0) circle (1pt);
\filldraw  (9,1) circle (1pt);
\filldraw  (9,2) circle (1pt);

\draw (9,0) -- (10,0);
\draw (9,1) -- (10,2);
\draw (9,2) -- (10,1);

\filldraw  (10,0) circle (2pt);
\filldraw  (10,1) circle (2pt);
\filldraw  (10,2) circle (2pt);

\draw (10.5,1) node[anchor=west] {$\cdots$};
\draw  (9,-0.5) circle (10pt);
\draw (8.6,-0.5) node[anchor=west] {$y_{20}$};
\draw (10.5,-0.5) node[anchor=west] {$\cdots$};
\draw (12,2.5) node[anchor=west] {$P_{5}P_{5}=P_{0}$};
\filldraw  (12,0) circle (2.0pt);
\filldraw  (12,1) circle (2.0pt);
\filldraw  (12,2) circle (2.0pt);

\draw (12,0) -- (13,2);
\draw (12,1) -- (13,1);
\draw (12,2) -- (13,0);

\filldraw  (13,0) circle (1pt);
\filldraw  (13,1) circle (1pt);
\filldraw  (13,2) circle (1pt);

\draw (13,0) -- (14,2);
\draw (13,1) -- (14,1);
\draw (13,2) -- (14,0);

\filldraw  (14,0) circle (2pt);
\filldraw  (14,1) circle (2pt);
\filldraw  (14,2) circle (2pt);

\draw  (13,-0.5) circle (10pt);
\draw (12.6,-0.5) node[anchor=west] {$y_{35}$};
\end{tikzpicture}
\caption{The case $n=3, L=2$.}   
\label{game_3}
\end{figure}
\end{center}
$$X_1=\begin{pmatrix}
    x_{11}^{(1)}  &  x_{12}^{(1)} &  x_{13}^{(1)} \\
    x_{21}^{(1)}  &  x_{22}^{(1)} &  x_{23}^{(1)} \\
    x_{31}^{(1)}  &  x_{32}^{(1)} &  x_{33}^{(1)} 
\end{pmatrix}, X_2=\begin{pmatrix}
    x_{11}^{(2)}  &  x_{12}^{(2)} &  x_{13}^{(2)} \\
    x_{21}^{(2)}  &  x_{22}^{(2)} &  x_{23}^{(2)} \\
    x_{31}^{(2)}  &  x_{32}^{(2)} &  x_{33}^{(2)} 
\end{pmatrix}.$$
For example, we have $15=2\cdot(3!)+3$ for $j=15$ so 
$$y_{15}=x_{1,\pi_2(1)}^{(1)}\cdot x_{2,\pi_2(2)}^{(1)}\cdot x_{2,\pi_2(3)}^{(1)}\cdot x_{1,\pi_3(1)}^{(2)}\cdot x_{2,\pi_3(2)}^{(2)}\cdot x_{2,\pi_3(3)}^{(2)}\cdot\prod\limits_{(\alpha,\beta)\neq(i,\pi_{a_{j,s}}(i))}\overline{x_{\alpha,\beta}^{(\gamma)}}=$$
$$=x_{1,2}^{(1)}\cdot x_{2,1}^{(1)}\cdot x_{3,3}^{(1)}\cdot x_{1,2}^{(2)}\cdot x_{2,3}^{(2)}\cdot x_{3,1}^{(2)}\cdot
\overline{x_{1,1}^{(1)}}\cdot\overline{x_{1,3}^{(1)}}\cdot
\overline{x_{2,2}^{(1)}}\cdot\overline{x_{2,3}^{(1)}}\cdot
\overline{x_{3,1}^{(1)}}\cdot\overline{x_{3,2}^{(1)}}\cdot
\overline{x_{1,1}^{(2)}}\cdot\overline{x_{1,3}^{(2)}}\cdot
\overline{x_{2,1}^{(2)}}\cdot\overline{x_{2,2}^{(2)}}\cdot
\overline{x_{3,2}^{(2)}}\cdot\overline{x_{3,3}^{(2)}}.$$
We can write similar expressions for the others $y_j.$

In this case, we have $$M_{11}=\{0,1\}, M_{12}=\{2,3\}, M_{13}=\{4,5\},$$ $$M_{21}=\{2,4\}, M_{22}=\{0,5\}, M_{23}=\{1,3\},$$ $$M_{31}=\{3,5\}, M_{32}=\{1,4\}, M_{33}=\{0,2\}$$
and
$$I_0=\{0,7,14,22,27,35\}, I_1=\{1,6,16,20,29,33\}, 
I_2=\{2,9,12,23,25,34\},$$
$$I_3=\{3,8,17,18,28,31\}, I_4=\{4,11,13,21,24,32\}, 
I_5=\{5,10,15,19,26,30\}.$$
Thus
$$z_{1,1}=\bigvee_{s\in I_0\cup I_1}y_s=
y_0\vee y_7\vee y_{14}\vee y_{22}\vee y_{27}\vee y_{35}\vee y_1\vee y_6\vee y_{16}\vee y_{20}\vee y_{29}\vee y_{33},$$
$$z_{1,2}=\bigvee_{s\in I_2\cup I_3}y_s=
y_2\vee y_9\vee y_{12}\vee y_{23}\vee y_{25}\vee y_{34}\vee y_3\vee y_8\vee y_{17}\vee y_{18}\vee y_{28}\vee y_{31},$$
$$z_{1,3}=\bigvee_{s\in I_4\cup I_5}y_s=
y_4\vee y_{11}\vee y_{13}\vee y_{21}\vee y_{24}\vee y_{32}\vee y_5\vee y_{10}\vee y_{15}\vee y_{19}\vee y_{26}\vee y_{30},$$
$$z_{2,1}=\bigvee_{s\in I_2\cup I_4}y_s=
y_2\vee y_9\vee y_{12}\vee y_{23}\vee y_{25}\vee y_{34}\vee y_4\vee y_{11}\vee y_{13}\vee y_{21}\vee y_{24}\vee y_{32},$$
$$z_{2,2}=\bigvee_{s\in I_0\cup I_5}y_s=
y_0\vee y_7\vee y_{14}\vee y_{22}\vee y_{27}\vee y_{35}\vee y_5\vee y_{10}\vee y_{15}\vee y_{19}\vee y_{26}\vee y_{30},$$
$$z_{2,3}=\bigvee_{s\in I_1\cup I_3}y_s=
y_1\vee y_6\vee y_{16}\vee y_{20}\vee y_{29}\vee y_{33}\vee y_3\vee y_8\vee y_{17}\vee y_{18}\vee y_{28}\vee y_{31},$$
$$z_{3,1}=\bigvee_{s\in I_3\cup I_5}y_s=
y_3\vee y_8\vee y_{17}\vee y_{18}\vee y_{28}\vee y_{31}\vee y_5\vee y_{10}\vee y_{15}\vee y_{19}\vee y_{26}\vee y_{30},$$
$$z_{3,2}=\bigvee_{s\in I_1\cup I_4}y_s=
y_1\vee y_6\vee y_{16}\vee y_{20}\vee y_{29}\vee y_{33}\vee y_4\vee y_{11}\vee y_{13}\vee y_{21}\vee y_{24}\vee y_{32},$$
$$z_{3,3}=\bigvee_{s\in I_0\cup I_2}y_s=
y_0\vee y_7\vee y_{14}\vee y_{22}\vee y_{27}\vee y_{35}\vee y_2\vee y_9\vee y_{12}\vee y_{23}\vee y_{25}\vee y_{34}.$$
\end{example}

\section{Deep neural network solution}
The calculation of the matrix $Z=X_1\cdot X_2\cdot\ldots\cdot X_L$ can be performed using a deep learning network, multiplying sequentially: $Y_1=X_1,$ $Y_{k}=Y_{k-1}\cdot X_{k},$ $(k=2,\ldots, L).$ Then $Z=Y_{L}.$ The network diagram is shown in the Fig. \ref{deep_neuro_net}.

Let $X_k=(x_{ij}^{(k)}),$ $Y_k=(y_{ij}^{(k)}).$ Then 
$$y_{ij}^{(k)}=y_{i1}^{(k-1)}\cdot x_{1j}^{(k)}\oplus y_{i2}^{(k-1)}\cdot x_{2j}^{(k)}\oplus\ldots\oplus y_{in}^{(k-1)}\cdot x_{nj}^{(k)}.$$ 
To calculate the entries of matrices, we use conjunction and addition modulo $2.$
\begin{center}
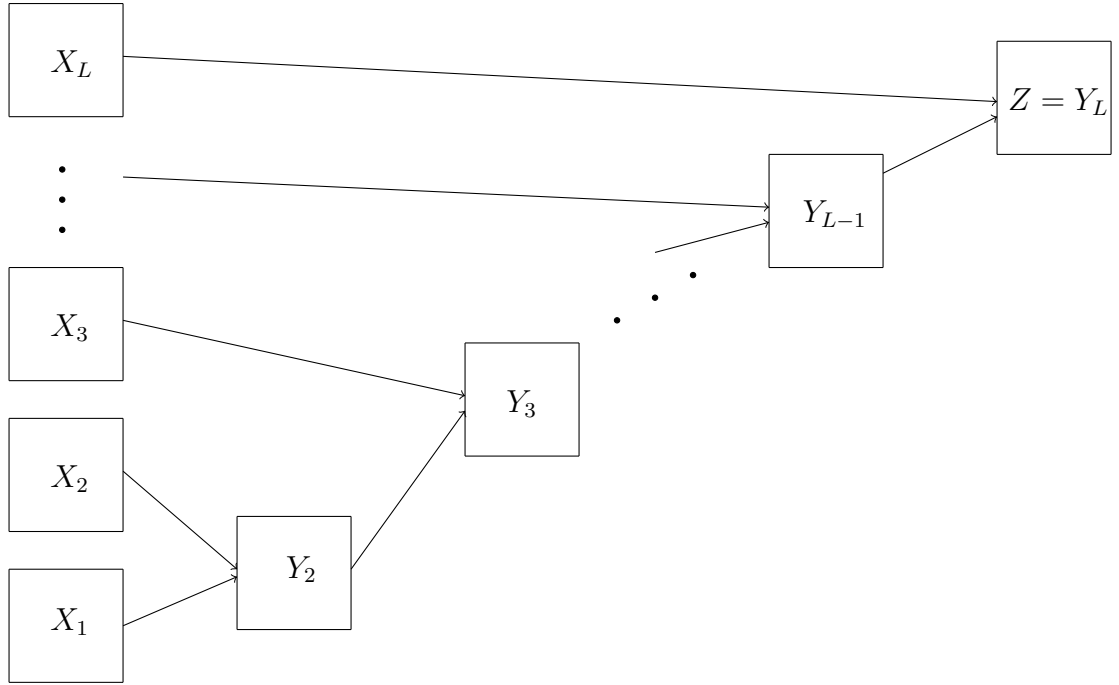
\begin{figure}[ht]
\centering  
\begin{tikzpicture}


\draw (0,0) -- (1.5,0);
\draw (0,1.5) -- (1.5,1.5);
\draw (0,0) -- (0,1.5);
\draw (1.5,0) -- (1.5,1.5);
\draw (0.4,0.8) node[anchor=west] {$X_1$};

\draw (0,2.0) -- (1.5,2.0);
\draw (0,3.5) -- (1.5,3.5);
\draw (0,2.0) -- (0,3.5);
\draw (1.5,2.0) -- (1.5,3.5);
\draw (0.4,2.7) node[anchor=west] {$X_2$};

\draw (0,4.0) -- (1.5,4.0);
\draw (0,5.5) -- (1.5,5.5);
\draw (0,4.0) -- (0,5.5);
\draw (1.5,4.0) -- (1.5,5.5);
\draw (0.4,4.7) node[anchor=west] {$X_3$};

\filldraw  (0.7,6.0) circle (1pt);
\filldraw  (0.7,6.4) circle (1pt);
\filldraw  (0.7,6.8) circle (1pt);

\draw (0,7.5) -- (1.5,7.5);
\draw (0,9.0) -- (1.5,9.0);
\draw (0,7.5) -- (0,9.0);
\draw (1.5,7.5) -- (1.5,9.0);
\draw (0.4,8.2) node[anchor=west] {$X_L$};

\draw (3,0.7) -- (4.5,0.7);
\draw (3,2.2) -- (4.5,2.2);
\draw (3,0.7) -- (3,2.2);
\draw (4.5,0.7) -- (4.5,2.2);
\draw (3.5,1.5) node[anchor=west] {$Y_2$};

\draw[->] (1.5,0.75) -- (3.0,1.4); 
\draw[->] (1.5,2.8) -- (3.0,1.5); 

\draw (6,3) -- (7.5,3);
\draw (6,4.5) -- (7.5,4.5);
\draw (6,3) -- (6,4.5);
\draw (7.5,3) -- (7.5,4.5);
\draw (6.4,3.7) node[anchor=west] {$Y_3$};

\draw[->] (4.5,1.5) -- (6.0,3.6); 
\draw[->] (1.5,4.8) -- (6.0,3.8); 

\draw (13,7) -- (14.5,7);
\draw (13,8.5) -- (14.5,8.5);
\draw (13,7) -- (13,8.5);
\draw (14.5,7) -- (14.5,8.5);
\draw (13.0,7.7) node[anchor=west] {$Z=Y_L$};

\draw (10,5.5) -- (11.5,5.5);
\draw (10,7.0) -- (11.5,7.0);
\draw (10,5.5) -- (10,7.0);
\draw (11.5,5.5) -- (11.5,7.0);
\draw (10.3,6.2) node[anchor=west] {$Y_{L-1}$};

\filldraw  (8.0,4.8) circle (1pt);
\filldraw  (8.5,5.1) circle (1pt);
\filldraw  (9.0,5.4) circle (1pt);

\draw[->] (11.5,6.75) -- (13.0,7.5); 
\draw[->] (1.5,8.3) -- (13.0,7.7); 
\draw[->] (1.5,6.7) -- (10.0,6.3); 
\draw[->] (8.5,5.7) -- (10.0,6.1); 

\end{tikzpicture}
\caption{A deep neural network diagram for simplified travel maze problem.}   
\label{deep_neuro_net}
\end{figure}
\end{center}
\begin{theorem}\label{DeepEstimate}
The constructed deep neural network has a depth of $L$, 
$$(2L-1)n^2$$ neurons, and $$2(L-1)n^3$$ connections between neurons.
\end{theorem}

\section{Neural network for $r$-bounded problem \label{Sec:bounded}}
A problem is called $r$-bounded ($0\leq r\leq n-1$) if the inequality $|\pi_{j_k}(i)-i|\leq r$ holds for all $i=1,\ldots,n;$ $k=1,\ldots,L.$ It means that the corresponding permutation matrices are banded matrices with the bandwidth $r + 1,$ i.e. $x_{ij}=0$ for $|i-j| \geq r + 1.$ 

Let $A$ be a banded matrix of the bandwidth $r + 1.$ The maximum number of nonzero entries in an arbitrary row of $A$ at most $2r + 1.$ The number of nonzero entries in $A$ at most $$N_r=n^2-(n-r)(n-r-1)=n(2r+1)-(r^2+r).$$
If $A$ and $B$ are banded matrices of the bandwidth $r + 1$ and $t + 1$, respectively, then the product $AB$ is a banded matrix of bandwidth $r + t + 1.$

\begin{theorem}  
For a $r$-bounded problem, there is a shallow neural network with a depth 
of $3$, $$ L\cdot N_r + n^2 + (n!)^L $$ neurons, and 
$$(L \cdot N_r + n^2 ) \cdot (n!)^L$$ connections between them.
\end{theorem}

\begin{theorem}  For a $r$-bounded problem, there is a deep neural network with a depth $L,$ 
 $$L\cdot N_r+\sum\limits_{i=2}^{L} N_{ir}\ \text{neurons if}\ Lr\leq n-1,$$
$$L\cdot N_r+\sum\limits_{i=2}^{[\frac{n-1}{r}]} N_{ir}+n^2\cdot\left(L-\left[\frac{n-1}{r}\right]\right)\ \text{neurons if}\ Lr> n-1,$$,   
$$2\sum\limits_{i=2}^{L} (ir+1)N_{ir} \ \text{connections between neurons if}\ Lr\leq n-1,$$
$$ \text{and } 2\sum\limits_{i=2}^{[\frac{n-1}{r}]} (ir+1)N_{ir}+ n^2\cdot(2r+n+1)\cdot\left(L-\left[\frac{n-1}{r}\right]\right)\ \text{connections if}\ Lr> n-1.$$
\end{theorem}

\section{Conclusion and outlook}

\begin{itemize}
\item Shallow neural network combined from elementary Rosenblatt's perceptrons can solve the travel maze problem, in  accordance with Rosenblatt's first theorem.
\item Complexity of the constructed solution of the travel maze problem by deep network is much smaller than for the solution provided by the shallow network (the main terms are   $2Ln^2$ versus $(n!)^L$ for the numbers of neurons and $2L^3$ versus $Ln^2 (n!)^L$ for the numbers of connections). 
\end{itemize}

The first result is important in the context of the widespread myth that elementary Rosenblatt's perceptrons have limited abilities and that Minsky and Papert revealed these limitations. This mythology has penetrated even into the encyclopedic literature \cite{History2015}.

Original Rosenblatt's perceptrons \cite{Rosenblatt1962} (Fig.~\ref{Fig:ElementaryPerceptron}) can solve any problem about classification of binary images and, after minor modification, even wider. This simple fact was proven in Rosenblatt's first theorem, and nobody criticised this theorem and proof.
The  universal representation property of shallow neural networks were studied in 1990s from different point of view, including approximation of real-valued functions \cite{Ito1998} and evaluation of upper bounds on rates of approximation \cite{Kurkova1998}. Elegant analysis of shallow neural networks involved infinite-dimensional hidden layers \cite{Kainen2009} and upper bounds were derived on the speed of decrease of approximation error as the number of network units increases. Abilities and limitations of shallow networks were reviewed recently in detail \cite{Kurkova2020}.

Of course, a single $R$-element can solve only linearly separable problems, and, obviously,
not all problems are linearly separable. Stating this trivial statement does not require any intellectual effort. Minsky and Papert \cite{Minsky1988} considered much more complex systems then a single linear threshold $R$-element. They studied the same elementary perceptrons that Rosenblatt did (Fig.~\ref{Fig:ElementaryPerceptron}) with one restriction: {\it receptive fields of $A$-elements are bounded}. These limitation may assume a sufficiently small diameter of the receptive field (the most common condition), or limited number of input connections of each $A$-neuron. Elementary perceptrons with such restriction have limited abilities: if we have only local information, then we cannot solve such a global problem as checking the connectivity of a set or the travel maze problem with one glance. We should integrate the local knowledge into global criterion using a sequence of steps. This intuitively clear statement was accurately formalised and proved for the parity problem by Minsky and Papert \cite{Minsky1988}.

Without restrictions, elementary perceptrons are omnipotent. In particular, they can solve the travel maze problem in the proposed form, but the complexity of solutions can be huge  (Theorem~\ref{ShallowEstimate}). On the contrary, the deep network solution (Theorem~\ref{DeepEstimate}) is much simpler and seems to be much more natural. It combined solution from the one-step permutations locally, step by step, whereas the shallow network operated by all possible global paths. Restriction of the possible paths of travel by bounded radius of a single step (Sec.~\ref{Sec:bounded}) does not change the situation qualitatively. (The restricted problem is simpler than the original one. This should not be confused with the possible network limitations, that complicate all problems.)

The second observation seems to be more important than the first one: the properly selected deep solutions can be much simpler than the  shallow solutions. In the contrast to the widely discussed huge deep structures and their surprising efficiency (see the detailed exposition of mathematics of deep learning in \cite{MathDep2021}) the relatively small but deep neural networks are non-surprisingly effective for solution of problems where local information should be integrated into global decision, like in the discussed version of the travel maze problem. These networks combine the benefits of the fine-grained parallel procession and the solutions of problems at a glance with the    possibility to emulate logic of sequential data analysis, when it is necessary. The important question in this context is: ``How deep should be the depth?'' \cite{GorbanDepth2020}. The answer depends on the problem.

The open question remains: are the complexity estimates sharp? How far are our solutions from the best ones? We do not expect that this problem has a simple solution because even for multiplication of $n \times n$ matrices no final solution has yet been found despite great efforts and significant progress (for the best of our knowledge, the latest improvement from $n^{2.37287}$ to $n^{2.37286}$ was achieved recently \cite{Alman2021}).
Another open question might attract attention: Analyse the  original geometric travel maze problem (see Fig.~\ref{Fig:MazeTravel} instead of its more algebraic simplification presented in Fig.~\ref{game}). It includes many non-trivial tasks, for example, convenient discrete representation of the possible paths with bounded curvature, lengths and ends, and constructive selection of $\epsilon$-networks in the space of such paths for preparing the input weights of of $A$-elements.

Complexity of functions computable by deep and shallow networks used for solution of classification problem were compared for the same complexity of networks \cite{Bianchini2014}. Complexity of functions was measured using topological invariants of the superlevel sets. The results seem to support the idea that deep networks can address more difficult problems with the same number of resources. 

The problem of effective parallelism pretends to be the central problem which is being
solved by the whole neuroinformatics \cite{Gorban1999}.  It has long been known that the efficiency of parallel computations increases slower than the number of processors.  There is a well known ``Minsky hypothesis'': efficiency of a parallel system increases (approximately) proportionally to logarithm of the number of processors; at least, it is a concave function.  Shallow neural networks pretend to solve all problems in one step, but the cost for that may be enormous number of resources. Deep networks make possible a trade-off between resources (number of elements) and the time needed to solve a problem, since they can combine the efficient parallelism of neural networks with elements of sequential reasoning. Therefore, neural networks can be a useful tool for solving the problem of efficient fine-grained parallel computing if we can answer the question: how deep should the depths be for different classes of problems. The case study presented in our note gives an example of significant increase of efficiency for a reasonable choice of depth.

\section*{Acknowledgments}
The project is supported by the Ministry of Science and Higher Education of the Russian Federation (Project No 075-15-2020-808).

\end{document}